 \documentclass[pmlr,twocolumn,10pt]{jmlr} 

\usepackage{booktabs}
\usepackage{siunitx}
\usepackage{svg}
\usepackage{amsmath}

\theorembodyfont{\upshape}
\theoremheaderfont{\scshape}
\theorempostheader{:}
\theoremsep{\newline}

\jmlrvolume{LEAVE UNSET}
\jmlryear{2023}
\jmlrsubmitted{LEAVE UNSET}
\jmlrpublished{LEAVE UNSET}
\jmlrworkshop{Machine Learning for Health (ML4H) 2023} 

 \title[HistoCap]{Automatic Report Generation for Histopathology images using pre-trained Vision Transformers}

\author{%
\Name{Saurav Sengupta}
\Email{ss4yd@virginia.edu}\\
\addr School of Data Science, University of Virginia, Charlottesville, USA
\AND
\AND
\Name{Donald Brown} \Email{deb@virginia.edu}\\
\addr School of Data Science, University of Virginia, Charlottesville, USA
}

\begin{document}

\maketitle

\begin{abstract}
Deep learning for histopathology has been successfully used for disease classification, image segmentation and more. However, combining image and text modalities using current state-of-the-art methods has been a challenge due to the high resolution of histopathology images. Automatic report generation for histopathology images is one such challenge. In this work, we show that using an existing pre-trained Vision Transformer in a two-step process of first using it to encode 4096x4096 sized patches of the Whole Slide Image (WSI) and then using it as the encoder and an LSTM decoder for report generation, we can build a fairly performant and portable report generation mechanism that takes into account the whole of the high resolution image, instead of just the patches. We are also able to use representations from an existing powerful pre-trained hierarchical vision transformer and show its usefulness in not just zero shot classification but also for report generation.
\end{abstract}





\section{Introduction}
\label{sec:intro}

High resolution histopathology slides are a rich resource of information that current deep learning methods are able to exploit for various use cases like disease classification, cell segmentation and outcome prediction. However, as the images are very high resolution, usually in the range of 150,000x150,000px, they often require non-trivial modifications to existing state-of-the-art (SOTA) deep learning architectures to be used successfully. The most common method for handling these high resolution images is to patch the bigger image into smaller sized images that can be fed into Convolutional Neural Networks. For example, in a classification setting, this often works as a multiple instance learning problem, where each patch is given the same overall image label. A potential drawback to this is that patching can lead to removal of overall context from the whole slide image (WSI) that the model might need to learn to make the correct decision, unless handled properly.

Automatic report generation for histopathology images is an area of existing research that also suffers from the need for modifying SOTA image captioning architectures to fit researchers needs. Image captioning for histopathology helps us combine two rich sources of information, that is, high resolution WSIs and associated diagnostic reports that describe features of the image. In clinical settings, automatic report generation has been successfully used for X-ray images and claim to reduce the burden for radiologists by assisting them in describing the image \citep{park2020feature}. Other use cases for automated image captioning in medical images can be image retrieval, as generated reports could be part of a searchable database, and encouraging standardized clinical ontologies by using words from a standard vocabulary to describe similar things. Therefore, automated image captioning for histopathology can be similarly useful for a wide variety of tasks that can assist physicians and radiologists in their tasks.

\section{Related Work}
Current research for histopathological image captioning focuses on CNN based encoder and Recurrent neural Network (RNN) based decoder architectures \citep{zhang2020evaluating, gamper2021multiple, tsuneki2022inference}. This is inspired by Show, attend and tell paper by \citet{xu2015show}, that in particular has the capability of using the attention mechanism to focus on certain areas of the image to generate captions. Attention values for explanation is an existing area of research as explained by \citet{haab2022attention} and is often a requirement to validate the correctness of the output in clinical settings where model reliability and interpretability is of the utmost importance.

Using Imagenet pre-trained Convolutional Neural Network (CNN) encoders to encode smaller sized patches of the high resolution WSI has been successfully used in a variety of ways to essentially reduce the size of large dimensional WSIs to smaller and computationally manageable representations. In recent years, \citet{chen2022scaling} have proposed a self-supervised Vision Transformer (ViT) based image representation learning mechanism called Hierarchical Image Pyramid Transformer (HIPT). The self-supervised pre-training leverages DINO (distillation with no labels) from \citet{caron2021emerging} at two levels, 256x256 sized patches and 4096x4096 sized patches. The authors show that this can then be leveraged for further downstream tasks like disease sub-typing and survival prediction, as the pre-trained ViT representations are now looking at the WSI at multiscale level. Our method uses these powerful Vision Transformers for encoding the whole WSI into a computationally manageable representation.

\citet{zhang2020evaluating} use a two-step process in which they first encode all patches of a WSI using a triplet loss based autoencoder and use the features from the bottleneck layer to cluster the patches into $k \in [1,2,3..7]$ clusters. In the second step they randomly sample the patches from each cluster, use a ImageNet pretrained ResNet-18 to extract $N$-dimensional features for the $k$-patches, then use attention pooling to reduce $k \times N$ dimensional feature vector to $1 \times N$ and then feed into a LSTM decoder to generate captions. Clustering using autoencoder based features is done to smartly sample different areas of the image with similar features instead of randomly sampling patches from the high resolution image, which can be often uninformative. However, this method is still lacking in terms of only having access to only $k$-patches of the Whole Slide Image instead of the entirety of the image while it generating the captions.

More recently, \citet{gamper2021multiple} describe the ARCH dataset which contains histopathology images extracted from textbooks and their associated descriptions as an analog to MS-COCO captions for histopathology, which they use for caption generation based pre-training task to generate self-supervised pre-trained encoder that when used for downstream tasks like multiple instance learning based classification shows promising results compared to other pre-trained encoders. But as noted in \citet{tsuneki2022inference}, these images, as they are curated from textbooks and research articles, can be of mixed quality, magnifications, and resolutions that while useful, does not really solve the problem for existing high resolution WSIs being generated in hospital systems everywhere.

In \citet{tsuneki2022inference}, the authors use high resolution WSIs from a Japanese hospital system and associated translated text reports with a vocabulary of size 277, for their automated captioning system. They use EfficientNetB3 \citep{tan2019efficientnet} and DenseNet121 \citep{huang2017densely} pretrained on ImageNet dataset and extract features from the penultimate layer for 300x300 patches extracted from the WSI. They then use global average pooling and 3x3 average pooling to reduce the feature sizes and feed them into an RNN based decoder for generating their captions.

In this paper, we use the same dataset as \citet{zhang2020evaluating} with their available train/val/test splits\footnote{https://github.com/zhangrenyuuchicago/PathCap}, specifically the same test data for comparison purposes and use pre-trained $\mathrm{ViT_{256}}$-16 representations from \citet{chen2022scaling} and an ViT based encoder and borrowing the LSTM based caption generator from \citet{xu2015show}. We show that we achieve comparable results on the same test set as used by \citet{zhang2020evaluating} and further introduce a method to utilize these powerful pre-trained transformers for a new downstream task, that is, automatic report generation for high resolution histopathology images.

\begin{figure*}[htbp]
\centering
\floatconts
  {fig:overview}
  {\caption{Method Overview}}
  {\includegraphics[width=0.9\linewidth]{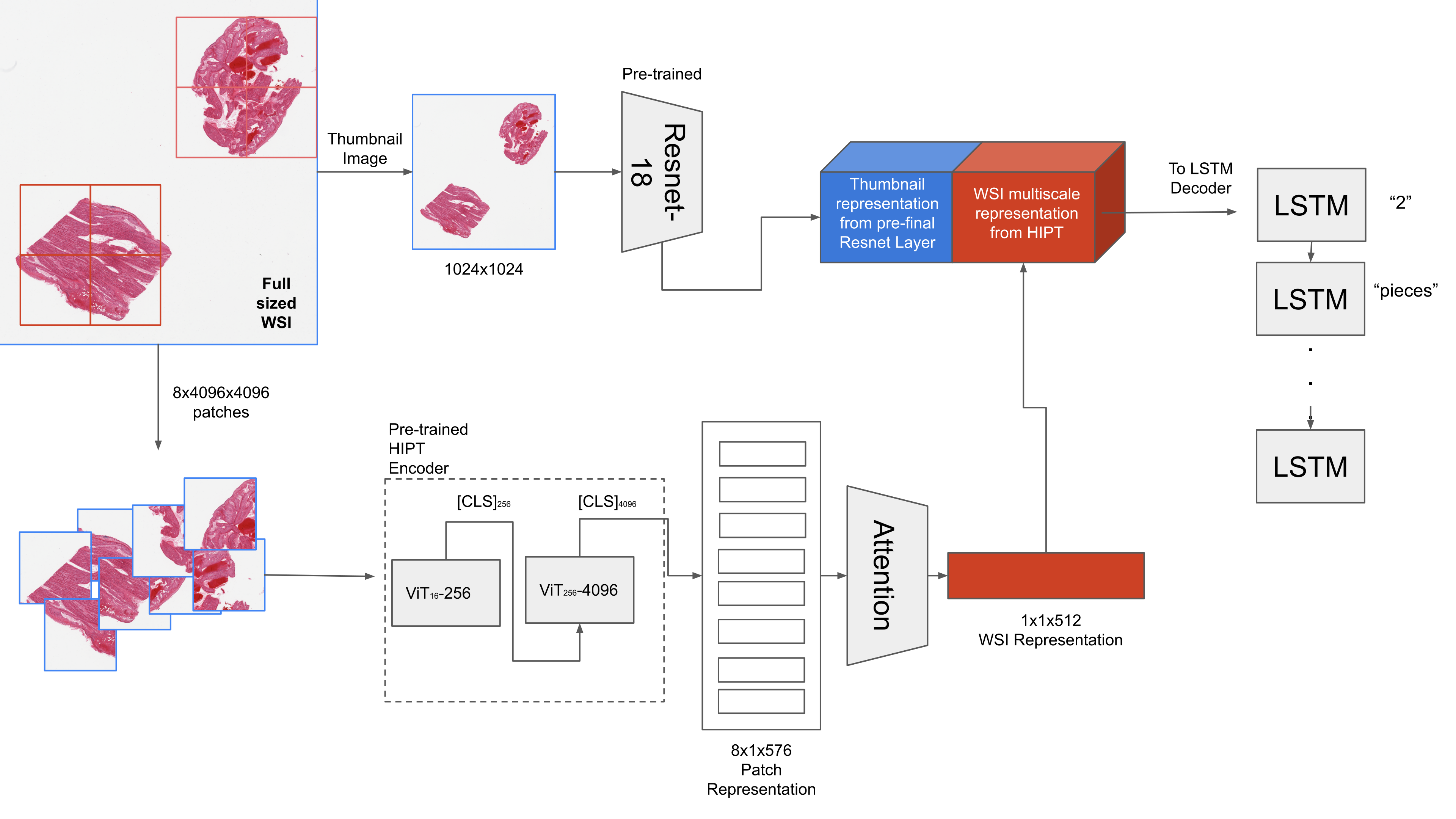}}
\end{figure*}

\section{Dataset}
We get our imaging and associated text data from the Genotype-Tissue Expression (GTEx) portal\footnote{https://www.gtexportal.org/home/histologyPage}, same as \citet{zhang2020evaluating} and use their train/val/test split such that we have 7989 training, 972 validation and 984 testing slides. For a complete and fair comparison, we also borrow the vocabulary, such that we have the same 971 tokens.

We do this because this is the only publicly available high resolution histology data with associated descriptions of each histology slide at the time of writing this paper that we could find. Data from \citet{tsuneki2022inference} comes in the form of 300x300 patches while publicly available\footnote{https://zenodo.org/record/6021442} with associated text descriptions, does not meet our criteria of working with high resolution Whole Slide Images, that is the most likely form of data available in health systems. And the ARCH dataset by \citet{gamper2021multiple}, as mentioned before, is derived from textbooks and research articles and again fails to meet this criteria.



\begin{table*}[htbp]
\floatconts
  {tab:results}
  {\caption{Caption Generation Results on test set}}%
  {%
    \begin{tabular}{|l|l|l|l|l|l|}
    \hline
    \abovestrut{2.2ex}\bfseries Model & \bfseries BLEU-1 & \bfseries BLEU-2 & \bfseries BLEU-3& \bfseries BLEU-4 &  \bfseries ROUGE\_L \\
    \hline
    \abovestrut{2.2ex}ResNet-18 (Random Init+Fine Tuning) & 0.3673 & 0.2688 & 0.1840 & 0.1189 & 0.4198\\
    ResNet-18 (Imagenet+Fine Tuning) & 0.3947 & 0.2895 &0.2009 & 0.1343& 0.4393\\
    Only ViT & 0.4023 & 0.2963 & 0.2094 & 0.1443& 0.4433\\
    \hline
    ResNet+ViT & \textbf{0.4116} & \textbf{0.3037} & \textbf{0.2147} & \textbf{0.1470}& \textbf{0.4480}\\
    PathCap \citep{zhang2020evaluating} & 0.4046 & 0.2986 & 0.2114 & 0.1455& 0.4290\\
    \hline
    \end{tabular}
    }
\end{table*}

\section{Method}

\subsection{HIPT architecture for WSI encoding}
\label{sec:wsienc}
To encode the whole WSI for caption generation, we use the HIPT (Hierarchical Image Pyramid Transformer) architecture as described in \citet{chen2022scaling}, where the authors first describe the self-supervised pre-training of multiscale Vision Transformers (ViT) using the DINO method for knowledge distillation pioneered by \citet{caron2021emerging} on 10,678 WSIs from The Genome Cancer Atlas (TCGA). They then describe using this pre-trained multiscale ViT for downstream tasks like slide level classification, survival prediction and further using the unique attention maps generated by the ViTs for finding morphological phenotypes. They show that a HIPT based WSI level encoder outperforms current SOTA in multiple instance learning for histopathology classification, that is, CLAM-SB \citep{lu2021data}.

Here, we explain the steps taken to encode the WSI using HIPT as that will inform our use later when we use it for report generation.

\begin{enumerate}
    
    \item Create $M$ 4096x4096 patches for each WSI, taking care that each patch contains more than 50\% tissue area.
    
    \item Initialize and freeze $\mathrm{ViT_{256}}$-16 and $\mathrm{ViT_{4096}}$-256 subnetworks. Here, in the notation $\mathrm{ViT_{L}}$-$l$, $L$ is the size of the patch and $l$ is the size of the $l \times l$ non overlapping tokens extracted from the $L \times L$ or $\textbf{x}_L$ image.
    
    \item For each $\textbf{x}_{4096}$ patch, first extract $\mathrm{[CLS]}_{256} \in \mathbb{R}^{1\times384}$ token vector from the $\mathrm{ViT_{256}}$-16, which is the summarizing token that encodes the 16x16 patch. And since we have 256 such 16x16 tokens per 256x256 patch, there are 256 such $\mathrm{[CLS]}_{256}$ tokens.
    
    \item The $\mathrm{[CLS]}_{256}$ token for each non overlapping $\textbf{x}_{256}$ patch from the $\textbf{x}_{4096}$ patch is extracted from the $\mathrm{ViT_{256}}$-16 and used to extract the $\mathrm{[CLS]}_{4096} \in \mathbb{R}^{1\times192}$ from $\mathrm{ViT_{4096}}$-256. This token now encodes the whole of $\textbf{x}_{4096}$ patch. Here, we also concatenate the mean of the 256 $\mathrm{[CLS]}_{256}$ to create $\mathrm{[CLS]}_{patch} \in \mathbb{R}^{1\times576}$ token, that is the final representation for each $\textbf{x}_{4096}$ patch.

    \item At the end, for each WSI we have a representation vector, $\mathcal{P} \in \mathbb{R}^{M\times576}$ consisting of $M$ $\mathrm{[CLS]}_{patch}$ tokens that can now be used for caption generation.
\end{enumerate}

\cite{chen2022scaling} generously provide their pre-trained weights for $\mathrm{ViT_{256}}$-16 and $\mathrm{ViT_{4096}}$-256 in a GitHub repository\footnote{https://github.com/mahmoodlab/HIPT/} that we utilize to generate multi-scale representations of our Whole Slide Image, that we can then use for caption generation.


\subsection{Caption Generation}\label{sec:capgen}

Here, we go over the steps for caption generation using a HIPT encoder also illustrated in Figure \ref{fig:overview}.

\begin{enumerate}
    \item Generate a low resolution (1024x1024) thumbnail image and $M$-4096x4096 sized patches at 20x magnification from all available WSIs.
    
    \item Use a pre-trained ResNet-18 to extract the thumbnail representation, $\mathcal{T} \in \mathbb{R}^{1\times512}$, and the pre-trained $\mathrm{ViT_{256}}$-16 and $\mathrm{ViT_{4096}}$-256 to extract patch based representation, $\mathcal{P} \in \mathbb{R}^{M\times576}$, as described in Section \ref{sec:wsienc}.
    
    \item Use a trainable Attention layer to create a weighted representation for the patch representations to generate a WSI level representation, $\mathcal{W} \in \mathbb{R}^{1\times576}$. This aggregated patch representation encodes the entire histology slide image. We also add a learnable projection layer $\phi$ to generate $\mathcal{W}' \in \mathbb{R}^{1\times512}$.
    
    \item Concatenate the thumbnail representation, $\mathcal{T} \in \mathbb{R}^{1\times512}$, and WSI level representation $\mathcal{W}' \in \mathbb{R}^{1\times512}$, such that we are using a multi-level representation for each slide, that is, at lower magnification with the thumbnail representation and higher magnification with the WSI representation using HIPT. The HIPT representations simultaneously encode information at a cellular level from $\mathrm{ViT_{256}}$-16, which is looking at extremely tiny 16x16 tokens at 20X magnification and at the tissue region level where $\mathrm{ViT_{4096}}$-256 is looking at 256x256 patches at the same magnification, which results in powerful representations.
    
    \item We feed this multi-level representation  for each WSI into the LSTM based decoder, similar to \citet{xu2015show}.
\end{enumerate}

There are a number of possible combinations of encoders, fine-tuning targets and representations that can be tweaked to find the best possible combination of all. For example, we can choose to initialize ResNet-18 randomly or use ImageNet pre-trained weights. We can also choose to fine-tune the ResNet-18 block or freeze it during training. In the following section, we detail our experiments.

\section{Experiments}
We first train a baseline model using only the thumbnail images for the WSIs and the same method as \citet{xu2015show}. We experiment first with a randomly initialized ResNet-18 and then with initializing it with Imagenet pre-trained weights. We then use only the ViT based representation for encoding the WSI, to understand the contribution of the ViT. And finally, we combine the thumbnail representations with the WSI level representations from the ViT based encoder and use that to feed the LSTM based decoder. 
For evaluating our model, we use the BLEU-4 score proposed by \citet{papineni2002bleu}. We restrict ourselves BLEU scores as it is being used by most researchers in automatic caption generation for histopathology \citep{tsuneki2022inference, zhang2020evaluating}.

We utilize PyTorch 1.12 and PyTorch Lightning 2.0.5 for our experiments (\href{https://github.com/ssen7/histo\_cap\_generation\_2}{code available here}). We use a batch size of 32, with mixed precision training and a learning rate scheduler that decays the learning rate by 0.8 when there is no increase in the validation BLEU-4 score for 8 consecutive epochs. We use Adam optimizer with encoder learning rate of 1e-4 and decoder learning rate of 4e-4. We also incorporate value based gradient clipping to tackle the exploding gradient problem. If the BLEU-4 score does not improve for 20 consecutive epochs, we stop the training and save the model with the best validation BLEU-4 score. All our models were trained on a single NVIDIA V100 GPU, and mixed precision training helped reduce training time from 10hrs to 2hrs. Both pre-trained vision transformers, that is, $\mathrm{ViT_{256}}$-16 and $\mathrm{ViT_{4096}}$-256 were frozen during training and only the attention layer and non-linear projection layer with ReLU activation were trained. The ResNet-18 was also unfrozen and fine-tuned during training.

We show our results in Table \ref{tab:results}. All BLEU scores and ROUGE\_L scores \citet{lin2004rouge} shown are an average of 5 runs except for results from \citet{zhang2020evaluating} which we borrow from the paper, as we are using the same test set. 

First, we can see that using ImageNet pre-training beats random initialization. We treat this as a sanity check for our method, since Imagenet pre-training in theory should outperform random initialization every time due to the powers of transfer learning. We can also see that using only ViT representations we beat the baseline results and are getting comparable results that are only slightly lower than that achieved by PathCap by \citet{zhang2020evaluating}. Combining ResNet-18 representations with ViT representations also slightly outperforms PathCap, which we credit to having an additional thumbnail representation from the ResNet-18. This method is effectively a multi-level way of encoding a WSI, as described in Section \ref{sec:capgen} and therefore performs comparably to PathCap which also looking at both the thumbnail and high-magnification views to perform the captioning task.



\section{Results and attention visualization}

Our method allows us to utilize both the attention map generation method described in \cite{xu2015show} and also use the trainable attention weights, $\alpha_i$, where $i \in [1,M]$ from the attention layer  to highlight the top $\textbf{x}_{4096}$ patches that the model weighted higher than the rest for generating the caption. This gives clinical professionals an additional method to validate that the model is looking at the right areas of the image for a certain caption being generated. Figure \ref{fig:attndiff} shows the caption generation performance and attention visualizations for the top performing ResNet+ViT model (BLEU-4=0.1544) for two different WSIs from our testing dataset.

While there has been much debate around whether attention values should be used for explanations as described by \citet{bibal2022attention}, \citet{haab2022attention} find that there is a higher chance of the interpretations being meaningful if the model itself is robust. The authors recommend using ensembling methods to improve those interpretations in case of failure modes where the model itself is good, but the attention values do not hold significant information. Therefore, to test this, in Figure \ref{fig:attnsame} we show the caption generation performance and attention maps for 3 top performing (in terms of test BLEU-4 scores) for the same WSI.

\begin{figure*}[htbp]
\floatconts
  {fig:attnsame}
  {\caption{Caption generation and Attention Visualizations for the same WSI, by 3 different models; (a) in each subfigure shows the actual WSI, (b) in each subfigure shows per word attention map generated on the thumbnail image, (c) in each subfigure shows the Top 4 patches as scored by the trainable Attention layer, (d) in each subfigure shows the predicted and actual captions associated with each WSI.}}
  {%
    \subfigure[WSI ID:GTEX-13FTW-1926 (Model 1;BLEU-4 Score=0.1544)]{\label{fig:attnsame1}%
      \includegraphics[width=0.6\linewidth]{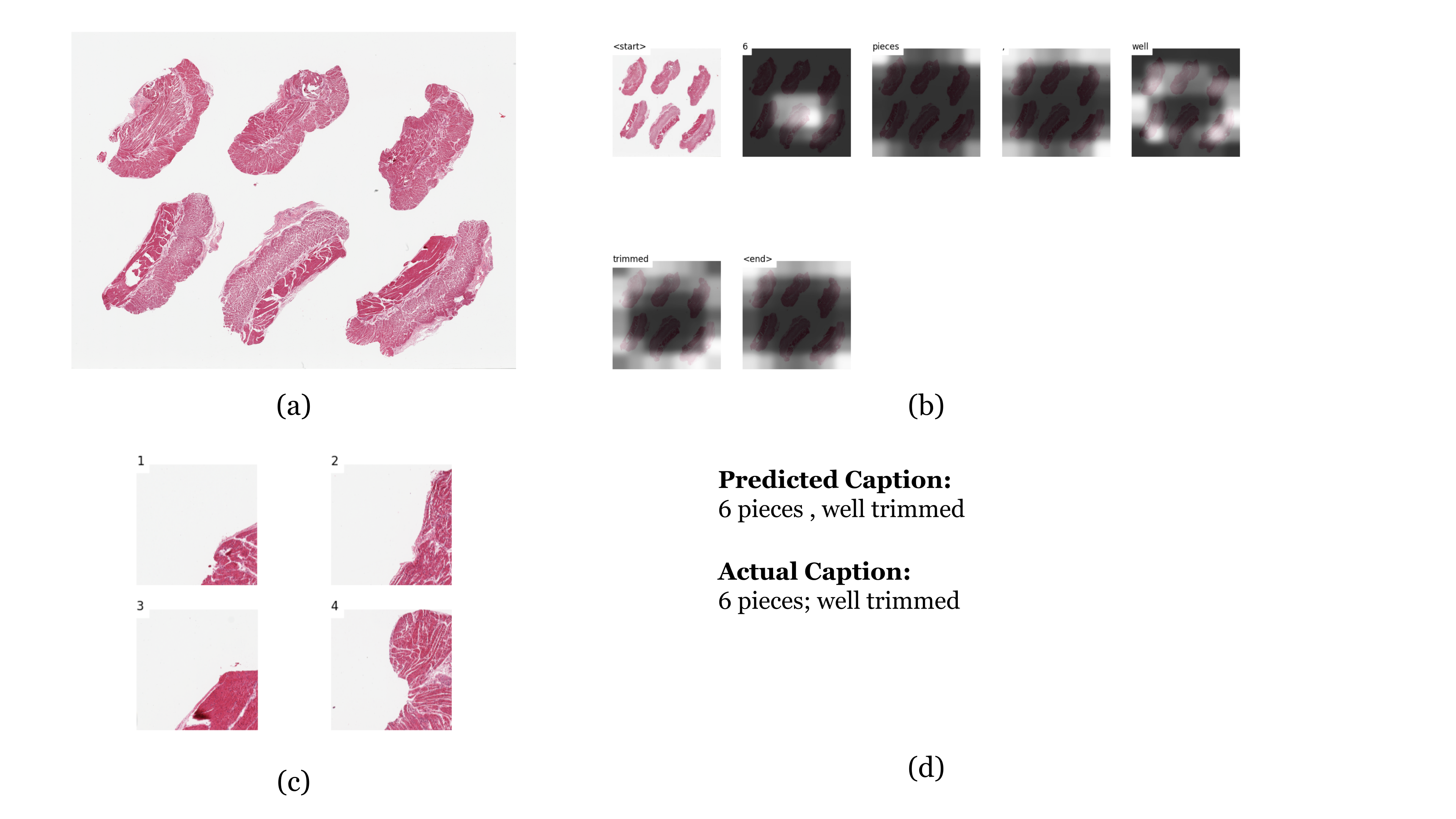}}
    \qquad
    \subfigure[WSI ID:GTEX-13FTW-1926 (Model 2;BLEU-4 Score=0.1475)]{\label{fig:attnsame2}%
      \includegraphics[width=0.6\linewidth]{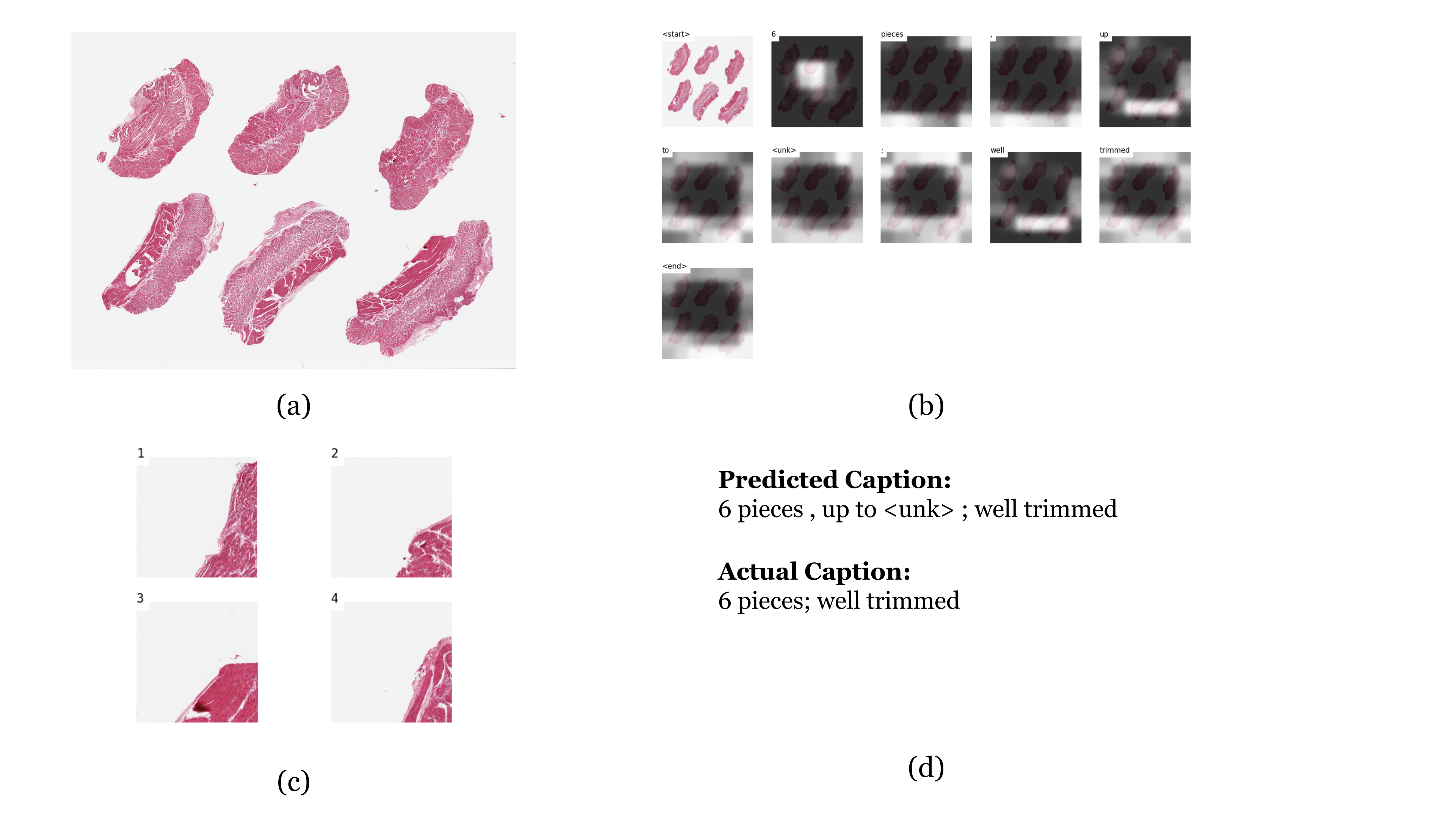}}
      \qquad
    \subfigure[WSI ID:GTEX-13FTW-1926 (Model 3;BLEU-4 Score=0.1493)]{\label{fig:attnsame3}%
      \includegraphics[width=0.6\linewidth]{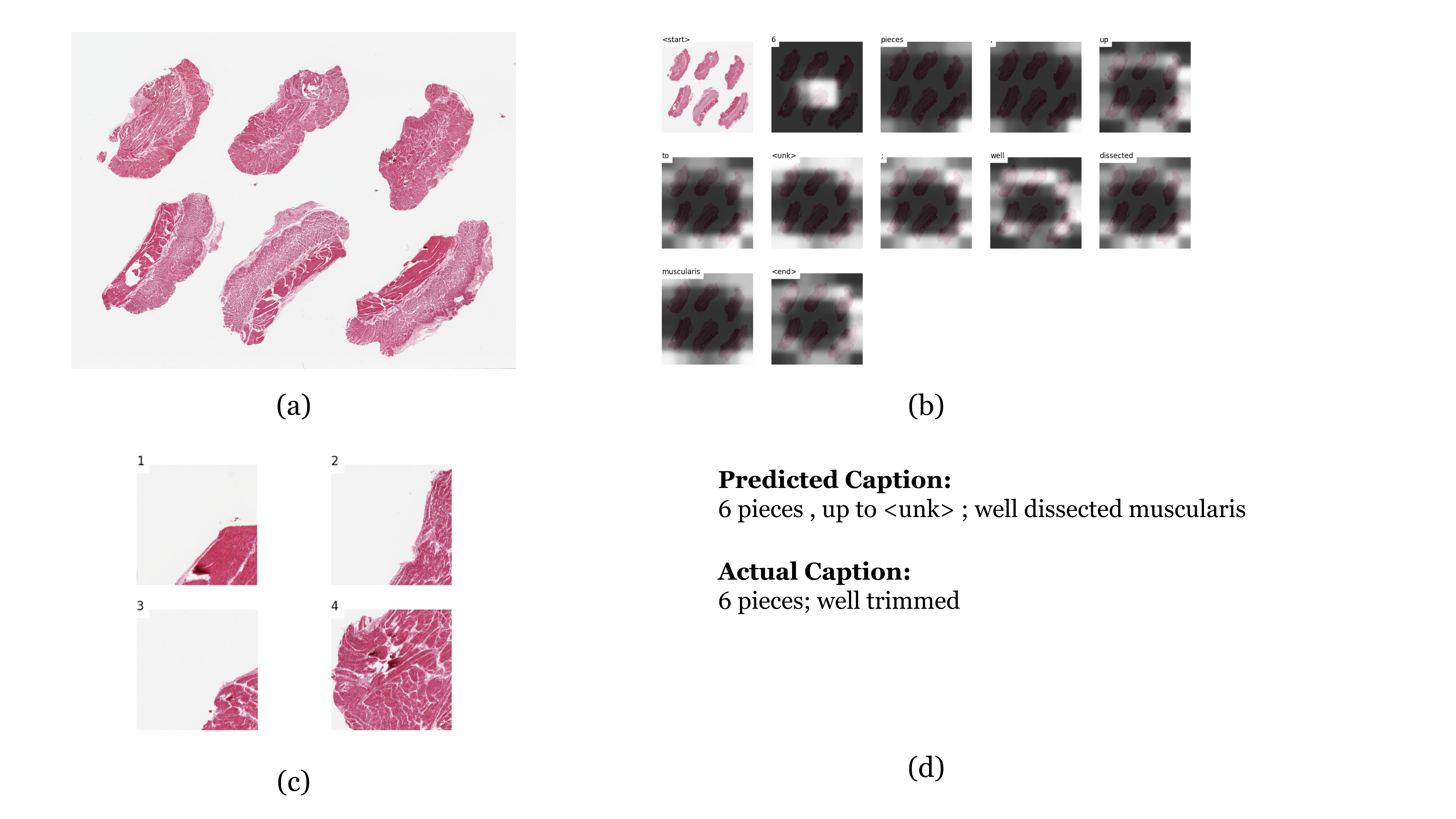}}
}

\end{figure*}

This shows that our caption generation model, when initialized differently, sometimes produces inconsistent captions and attention maps. One potential solution for this, motivated by \citet{haab2022attention}, can be doing an ensemble of all trained and performant models and then having a voting mechanism, such that only those captions that are being generated multiple times by different caption generators are accepted as the final caption, while the less occurring captions are not used. Our results show that multiple trained caption generators can produce similar captions that follow the reference caption wording closely, see Figure \ref{fig:attnsame1} and \ref{fig:attnsame2}, while the caption in \ref{fig:attnsame3} is completely different from the reference caption, which suggests such ensembling could potentially work. To potentially calculate how similar each generated caption is, we can embed the captions using a powerful Bidirectional Encoder Representations from Transformers (BERT) that are specifically trained on clinical and healthcare data such as one by \citep{lee2020biobert} and cluster the most similar ones based on cosine distance.


\begin{figure*}[htbp]
\floatconts
  {fig:attndiff}
  {\caption{Caption generation and Attention Visualizations for 2 different WSIs using the top performing model; (a) in each subfigure shows the actual WSI, (b) in each subfigure shows per word attention map generated on the thumbnail image, (c) in each subfigure shows the Top 4 patches as scored by the trainable Attention layer, (d) in each subfigure shows the predicted and actual captions associated with each WSI.}}
  {%
    \subfigure[Tissue Sample ID: GTEX-13FLV-0326]{\label{fig:attndiff1}%
      \includegraphics[width=0.9\linewidth]{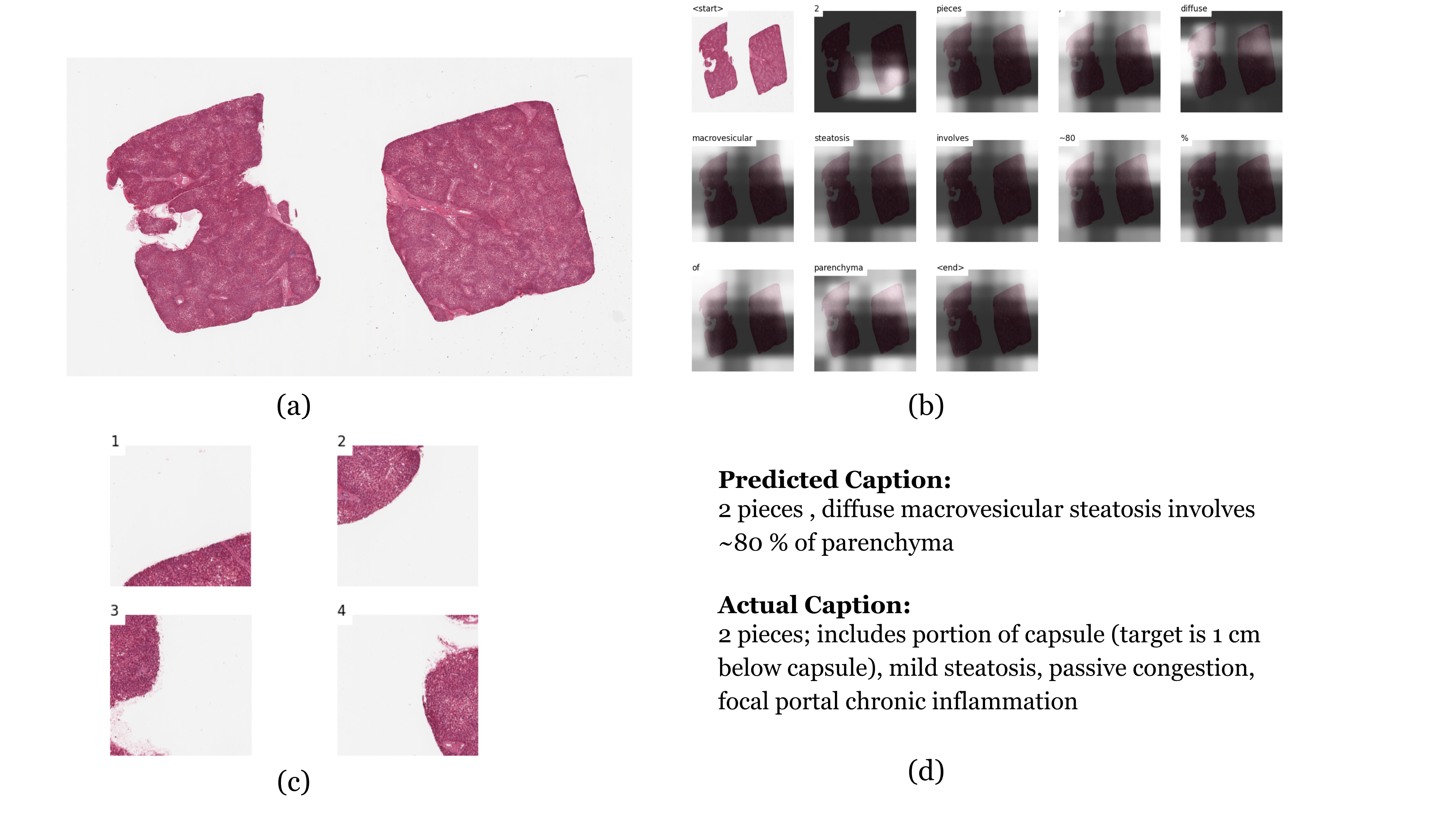}}
    \qquad
    \subfigure[Tissue Sample ID: GTEX-13NYS-0126]{\label{fig:attndiff2}%
      \includegraphics[width=0.9\linewidth]{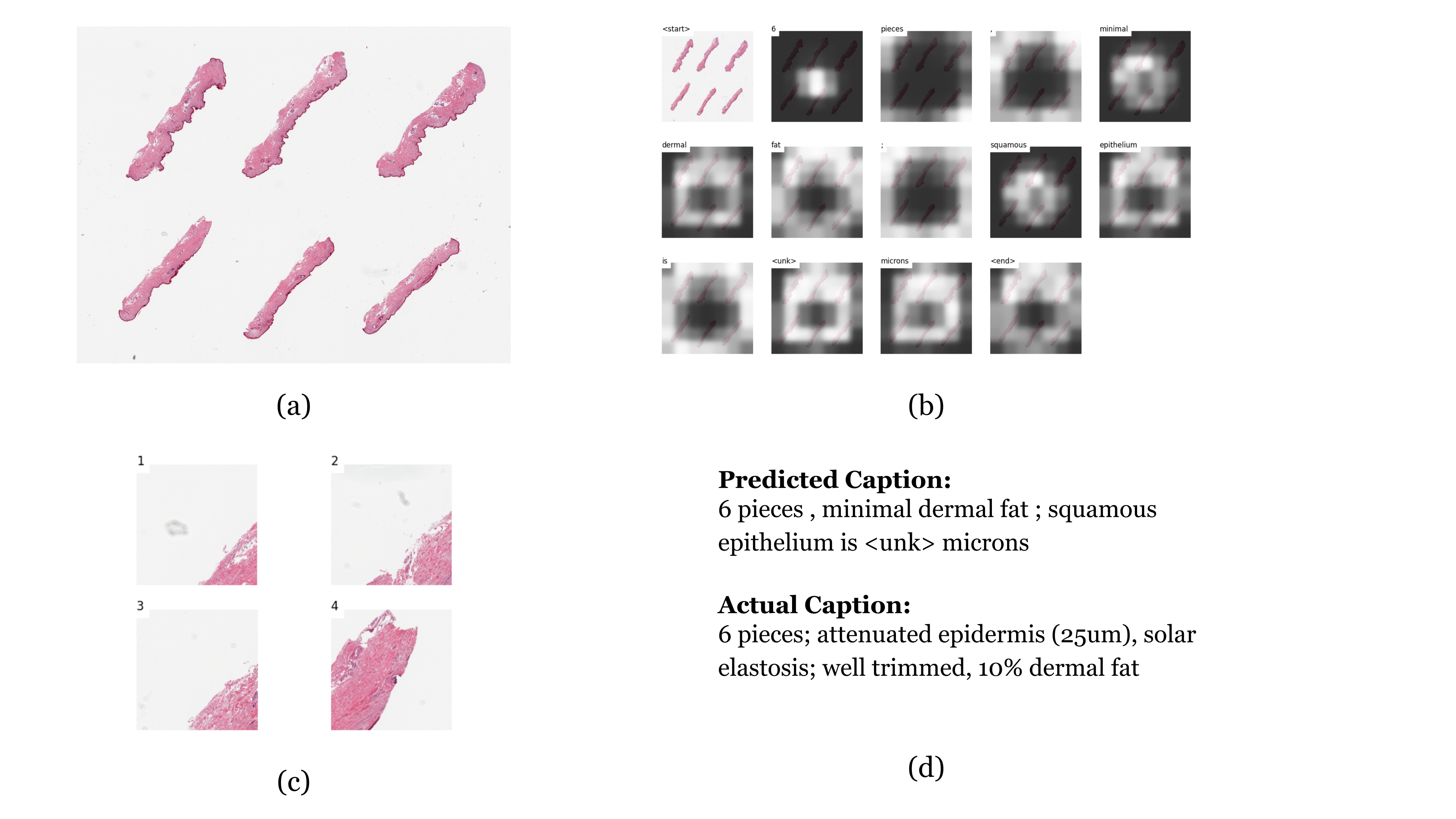}}
  }
\end{figure*}

\section{Discussion}

In this paper, we show that powerful pre-trained ViT based representations could be used to encode a very high resolution histology image slide for another downstream task, that is, successful automatic report generation. These pre-trained representations are not only useful for zero-shot learning, classification but can be adapted for different tasks as we have shown here. We have also validated that self supervised pre-training is helpful, as it lessens the burden of training data hungry models from scratch every time. Most previous methods for caption generation in histopathology required having a method to encode these high resolution images into manageable representations that could then be used for model learning, which can now successfully be done by powerful self-supervised pre-trained encoders and used for any number of tasks, in this case, caption generation.

Additionally, we can look at caption generation as a self-supervised objective that allows us to combine images and text based data. Data like radiology and histopathology slides and associated text reports are stored across hospital systems and can be used for self-supervised learning and do not require trained clinical professionals to assign labels that is often necessary for most supervised learning tasks. Indeed, \citet{lu2023visual} and \citet{gamper2021multiple} propose similar objectives, that is, combining vision and language data to pre-train powerful encoders for histopathology that can then be used for zero-shot learning and produce promising results. 

In the same vein, we can look at the caption generation as a way to incorporate text based data into these already powerful representations generated by the vision transformers trained by \cite{chen2022scaling}. We present, in this paper, an automatic caption generation mechanism for histopathology that we believe is easily trainable given new data, incorporates powerful multi-level representations and can potentially generate richer representations by combining image and text based data, that can improve performance on downstream tasks that previously would have required expensive labels to train performant models.

\textbf{Limitations}: This paper is limited by the data we are using, which are usually single sentence text descriptions of high resolution histology slides. Real world reports would definitely have longer sequences of words, and a larger vocabulary that would necessitate more complex decoders than LSTM based ones. Also, in healthcare, there can be words that could potentially have the same meaning but are a function of standards and practices followed by clinical professionals in different regions and health systems, which potentially hampers creation of a universally performant image captioning model.  We have also not tested our claims of the representations being richer due to incorporating text based data, which we take as future work. We are also far behind state-of-the-art in image caption generation for regular images, as the reported BLEU scores show, and there is room for improvement in this area.

\acks{Authors are funded by the NIH award UL1TR003015 for the integrated Translational Health Research Institute of Virginia.}

\bibliography{jmlr-sample}

\appendix





\end{document}